\documentclass[	
	paper = a4,
	abstract = true
]{scrartcl}
\usepackage[dvips]{graphicx}
\usepackage[utf8]{inputenc}
\usepackage{amssymb,amsmath,array}

\usepackage{xspace}
\usepackage{url}
\usepackage{booktabs}
\usepackage{multirow}
\usepackage{mathtools}
\usepackage{bm}
\usepackage{microtype}
\usepackage{times}
\let\vec\bm
\newcommand{\norm}[1]{\left\lVert#1\right\rVert}
\newcommand{\tp}[0]{\top}%
\newcommand{\lab}[0]{\tau}
\newcommand{\X}{\ensuremath{\mathcal{X}}\xspace}
\newcommand{\bbN}[0]{\ensuremath{\mathbb{N}}\xspace}
\newcommand{\bbRnn}[0]{\ensuremath{\mathbb{R}_{\geq0}}\xspace}
\newcommand{\Assign}{\ensuremath{\mathfrak{B}}\xspace}
\newcommand{\Hilb}{\ensuremath{\mathcal{H}}\xspace}
\DeclarePairedDelimiter\multiset{\lbrace\!\!\lbrace}{\rbrace\!\!\rbrace}%

\begin{document}
\title{Deep Weisfeiler-Lehman Assignment Kernels 
via Multiple Kernel Learning\footnote{Preprint of the ESANN 2019 paper~\cite{Kriege2019}.}}

\author{Nils M.~Kriege%
\thanks{This work was supported by the German Research Foundation (DFG) within the 
SFB 876 ``Providing Information by Resource-Constrained Data Analysis'', project A6 
``Resource-efficient Graph Mining''.}
\vspace{.3cm}\\
Department of Computer Science, TU Dortmund University\\
Otto-Hahn-Str.~14, 44227 Dortmund, Germany}

\date{\vspace{-1cm}}

\maketitle

\begin{abstract}
Kernels for structured data are commonly obtained by decomposing objects into their parts
and adding up the similarities between all pairs of parts measured by a base kernel.
Assignment kernels are based on an optimal bijection between the parts and have 
proven to be an effective alternative to the established convolution kernels.
We explore how the base kernel can be learned as part
of the classification problem. We build on the theory of valid assignment kernels
derived from hierarchies defined on the parts. We show that the weights of this 
hierarchy can be optimized via multiple kernel learning. 
We apply this result to learn vertex similarities for the Weisfeiler-Lehman 
optimal assignment kernel for graph classification. We present first experimental 
results which demonstrate the feasibility and effectiveness of the approach.
\end{abstract}

\section{Introduction}
Graphs are a versatile concept used to represent structured data in many domains
such as chem- and bioinformatics, or social network analysis.
Graph kernels have become an established and widely-used technique for solving
classification tasks on graphs. 
In the past 15 years a large number of graph kernels have been proposed.
One of the most successful in practice is the \emph{Weisfeiler-Lehman 
subtree kernel}~\cite{Shervashidze2011}, which is based on the color refinement 
heuristic for graph isomorphism testing. 
Recently, the \emph{Weisfeiler-Lehman optimal assignment kernel}~\cite{Kriege2016b}
was proposed, which also uses color refinement, but derives the kernel from an 
optimal bijection of the vertices instead of summing over all pairs of vertices.
In classification experiments, this approach yields higher accuracy scores
than the Weisfeiler-Lehman subtree kernel for many data sets.
More recently, several deep learning approaches to graph classification based on 
neural networks have emerged. These methods construct a vector representation for 
each vertex by iteratively applying a neighborhood aggregation function with 
learned weights. 
Most of them 
fit into the \emph{neural message passing framework} proposed in~\cite{Gil+2017}
and show promising results on several graph classification 
benchmarks~\cite{Yin+2018}.
Compared to these approaches, kernel methods are considered ``shallow'' since they do
not learn a representation by means of weights organized in a hierarchical manner.
This also is the case for the deep graph kernels proposed in~\cite{Yanardag2015a}, 
which support weights between graph features, but do not learn them 
end-to-end.
However, there are kernel methods that can justifiably be described as 
deep~\cite{Shawe-Taylor2014}. This is the case for multiple kernel learning, 
which was, for example, used to combine base kernels organized in a hierarchy 
according to their level of abstraction~\cite{Donini2016}.
Moreover, it was used to alleviate the diagonal dominance problem of convolution
kernels for graphs~\cite{Aiolli2015a}.

We propose deep assignment kernels for structured data which learn the 
base kernel on substructures as part of the training. To this end, 
we consider base kernels represented by a hierarchy 
for which the weights are obtained via multiple kernel learning.
Building on this, we propose a deep method 
for graph classification based on the Weisfeiler-Lehman method.
The feasibility of the approach is demonstrated experimentally.

\section{Basic Techniques}
We introduce the fundamentals and key techniques relevant for our contribution
in the following.
A \emph{kernel} on a set \X is a function $k : \X \times \X \to \mathbb{R}$ such 
that there is a real Hilbert space $\Hilb$ and a mapping $\phi : \X \to \Hilb$ 
with $k(x,y) = \langle \phi(x), \phi(y) \rangle$ for all $x,y$ in \X, 
where $\langle \cdot, \cdot \rangle$ denotes the inner product of $\Hilb$. 
We consider simple undirected graphs $G=(V,E)$, where $V(G) = V$ is the set of 
\emph{vertices} and $E(G)=E$ the set of \emph{edges}. An edge $\{u,v\}$ is for
short denoted by $uv$ or $vu$, where both refer to the same edge.
A graph with a unique path between any two vertices is a \emph{tree}.
A \emph{rooted tree} is a tree $T$ with a distinguished vertex $r \in V(T)$ 
called \emph{root}.

\subsection{Optimal Assignment Kernels}\label{sec:oa}
A common approach to compare two graphs is to construct an assignment 
between their vertices
that maximizes the structural overlap or agreement of vertex attributes.
This principle was proposed to obtain graph kernels, where the similarity 
between two vertices is determined by a base kernel~\cite{Frohlich2005}.
However, it was shown that the resulting similarity measure is not a valid kernel
in general~\cite{Vert2008}.
More recently, it was proven that for a specific class of base kernels, the 
similarity derived from optimal assignments is guaranteed to be a valid 
kernel~\cite{Kriege2016b}.

Let $[\X]^n$ denote the set of all $n$-element subsets of a set $\X$ and 
$\Assign(X,Y)$ the set of all bijections between $X,Y$ in $[\X]^n$ for 
$n \in \bbN$. 
The \emph{optimal assignment kernel} $K_\Assign^k$ on $[\X]^n$ is defined as
\begin{equation}\label{eq:assignment_kernel}
 K_\Assign^k(X,Y) = \max_{B \in \Assign(X,Y)} \sum_{(x,y) \in B} k(x,y),
\end{equation}
where $k$ is a \emph{base kernel} on \X. For the application to sets of different
cardinality, the smaller set can be augmented by dummy elements with no effect on
the solution value.
Given that the base kernel $k$ satisfies the \emph{strong kernel property}, i.e., 
$k(x,y) \geq \min\{k(x,z), k(z,y)\}$ for all $x,y,z \in \X$, the function 
$K_\Assign^k$ is a valid kernel~\cite{Kriege2016b}.
Strong kernels are equivalent to kernels obtained from a hierarchical partition 
of their domain. Formally, let $T$ be a rooted tree such that the leaves of $T$ 
are the elements of \X and $\omega : V(T) \to \bbRnn$ a weight function. 
We refer to the tuple $(T,\omega)$ as a \emph{hierarchy}. 
A hierarchy on \X induces a similarity $k(x,y)$ for $x,y \in \X$  as follows. 
For $v \in V(T)$ let $P(v) \subseteq V(T)$ denote the vertices in $T$ on
the path from $v$ to the root $r$.
Then the similarity between 
$x, y \in \X$ is $$k(x,y) = \sum_{v \in P(x) \cap P(y)} \omega(v).$$
For every strong kernel $k$ there is a hierarchy that induces $k$ and, vice versa,
every hierarchy induces a strong kernel~\cite{Kriege2016b}.

The optimal assignment kernel of Eq.~\eqref{eq:assignment_kernel}
can be computed in linear time from the hierarchy $(T, \omega)$ of the base kernel 
$k$ by histogram intersection as follows. For a node $v \in V(T)$ and a set 
$X \subseteq \X$, let $X_v$ denote the subset of $X$ that is contained in the 
subtree rooted at $v$. 
Then the optimal assignment kernel is
\begin{equation}\label{eq:oa}
 K_\Assign^k(X,Y) =  \sum_{\mathclap{v \in V(T)}} \min\{|X_v|, |Y_v|\} \cdot \omega(v),
\end{equation}
which can be seen as the histogram intersection kernel for appropriately defined 
histograms representing the sets $X$ and $Y$ under the strong base kernel $k$~\cite{Kriege2016b}.

\subsection{Weisfeiler-Lehman Optimal Assignment Kernels}\label{sec:wloa}
Color refinement, also known as 1-dimensional Weisfeiler-Lehman refinement or 
na\"ive vertex classification, is a classical heuristic for graph isomorphism 
testing. It iteratively refines partitions of the vertices of a graph, where the 
vertices in the same cell are said to have the same color.
In each iteration two vertices with the same color obtain different new colors
if their neighborhoods differ w.r.t.\@ the current coloring.
More recently, the approach was used to obtain kernels between graphs~\cite{Shervashidze2011}.
For the Weisfeiler-Lehman subtree kernel a graph is 
represented by a feature vector, where each component is associated with a color 
and counts the number of vertices in the graph having that color in one iteration.
The Weisfeiler-Lehman subtree kernel is the dot product of such feature vectors.

We give a clear mathematical formulation of the procedure. 
Given a parameter $h$ and a graph $G$ with initial colors $\lab_0$,
a sequence $(\lab_1,\dots,\lab_h)$ of refined  colors is computed,
where $\lab_i$ is obtained from $\lab_{i-1}$ by the following procedure.
For every vertex $v\in V(G)$, sort the multiset of colors 
$\multiset{\lab_{i-1}(u) :\, vu \in E(G)}$ to obtain a unique sequence of colors 
and add $\lab_{i-1}(v)$ as first element. Assign a new color $\lab_i(v)$ to every 
vertex $v$ by employing an injective mapping from color sequences to new colors. 
It was observed in~\cite{Kriege2016b} that color refinement applied to a set of 
graphs under the same injective mapping yields a hierarchy on the 
vertices. This hierarchy with a uniform weight function induces
the strong base kernel 
\begin{equation}\label{eq:wloa:vsim}
k(v,v') = \sum_{i=0}^h k_\delta(\tau_i(v),\tau_i(v'))
\end{equation}
on the vertices, where $k_\delta$ denotes the Dirac kernel.
This kernel measures the number of iteration required to assign different colors
to the vertices and reflects the extent to which the vertices have a structurally 
similar neighborhood. 
The optimal assignment kernel with this base kernel is referred to as 
\emph{Weisfeiler-Lehman optimal assignment kernel} and was shown to achieve
better accuracy results in many classification experiments than the 
Weisfeiler-Lehman subtree kernel.

\subsection{Multiple Kernel Learning}
Multiple kernel learning refers to machine learning techniques that extend
classical kernel-based support vector machines to use multiple (heterogeneous) 
kernels, which are combined using coefficients learned as part of the training.
One such approach is EasyMKL, which is scalable to a large number of kernels 
and can be used to obtain a data-driven feature weighting~\cite{Aiolli2015}.
EasyMKL combines the kernels $k_i$, $i \in \{1,\dots, R\}$, to a kernel 
$k(x,y) = \sum_{i=1}^R \alpha_i k_i(x,y)$ by learning the coefficients 
$\alpha_i \geq 0$. 
This is achieved by optimizing the problem
\begin{equation}\label{eq:mkl}
\max_{\vec{\alpha} : \norm{\vec{\alpha}}=1} \min_{\vec{\gamma} \in \varGamma}
(1-\lambda)\vec{\gamma}^\tp \vec{Y} \left(\sum_{i=0}^R  \alpha_i \vec{K}_i \right)
\vec{Y}\vec{\gamma} + \lambda \norm{\vec{\gamma}}_2^2,
\end{equation}
where $\vec{K}_i$ is the $l \times l$ kernel matrix obtained by applying $k_i$ 
on the training set of cardinality $l$, $\vec{Y}$ the diagonal matrix with 
$y_{i,i}$ the class label of the $i$th training example and $\lambda$ a 
hyperparameter for regularization.
The set $\varGamma$ is the domain of probability distributions 
$\vec{\gamma} \in \bbRnn^l$ defined over the sets of positive and negative
training examples.

\section{Deep Assignment Kernels}
We investigate how the base kernel used to compare the parts can be 
learned as part of the classification problem when comparing structured objects 
with an assignment kernel.
There are several basic difficulties with this general approach.
Since we derive a similarity measure from a combinatorial problem, it is 
not clear how changing the base kernel will effect the value of the optimal 
assignment. In particular, small changes in the base kernel value may lead to
entirely different optimal assignments.
Solving a single assignment problem takes cubic time in general, which is not
feasible for large instances.
Moreover, the value of assignments does not yield a valid kernel in general.

Therefore, we consider a highly restricted class of base kernels. 
We learn the base kernel from the class of kernels that are induced by the same 
tree $T$. Every base kernel in this class is uniquely defined by $T$ and the 
weight function $\omega$, which we would like to learn as part of the training.
Following the results summarized in Sec.~\ref{sec:oa} and considering
Eq.~\eqref{eq:oa} it becomes apparent that this can be achieved by multiple
kernel learning.
For every node $v \in V(T)$ we consider the kernel $k_v(X,Y) = \min\{|X_v|,|Y_v|\}$.
Solving Eq.~\eqref{eq:mkl} yields coefficients $\alpha_v$ with $\norm{\vec{\alpha}}=1$
that can be interpreted as learned weights $\omega(v) = \alpha_v$ for the tree 
$T$ to form a hierarchy.
In this way, we hope to obtain strong kernels that are adaptive to the specific 
learning task. However, the adaption is possible to a limited extent only: 
The tree $T$ determines a set of optimal solutions to the assignment problem. 
These solution will remain optimal under all learned weight functions, though 
their value may change. In case of $\alpha_v=0$ an equivalent hierarchy is 
obtained by removing the node $v$ from the tree and attaching its children to 
the parent of $v$. In this case the set of optimal solutions may be a superset
of the optimal solutions obtained for the tree with uniform weights.

\subsection{Deep Weisfeiler-Lehman Assignment Kernels}
We apply the observations stated above to the Weisfeiler-Lehman optimal assignment 
kernel, which is based on the hierarchy generated by color refinement as 
detailed in Sec.~\ref{sec:wloa}.
The hierarchy induces a similarity between the vertices, such that with each 
level the extent of the considered neighborhood increases.
In its original version uniform weights where used that induce the vertex 
similarity stated in Eq.~\eqref{eq:wloa:vsim}. Introducing weights as above,
we obtain the vertex similarity
\begin{equation}
k(v,v') = \sum_{i=0}^h \omega(\tau_i(v)) \cdot k_\delta(\tau_i(v),\tau_i(v'))
\end{equation}
as base kernel, where $\omega$ is a weight function for the colors of the 
refinement process. 
We refer to the assignment kernel with this base kernel as 
\emph{Deep Weisfeiler-Lehman assignment kernel}.
Please note that the value of the parameter $h$ is typically determined by
a grid search in a costly cross-validation process. Our approach is less 
dependent on the choice of this parameter since too specific features are
down weighted automatically.

\subsection{Feature and Weight Grouping}
Depending on the strong kernel and the data set, the representing hierarchy 
and therefore also the number of weights may be very large.
This is, for example, the case for the Deep Weisfeiler-Lehman assignment kernel,
since the color refinement process effectively distinguishes vertices with
different neighborhoods.
A very large number of weights slows down the training and may lead to
overfitting. To control the number of weights, we group the nodes of the 
hierarchy using a clustering algorithm.
To this end, we represent each node $v$ in the hierarchy by the data point 
$(c_1^v, c_2^v, \dots, c_n^v)$, where $n$ is the size of the data set and
$c_i^v$ is the number of elements of the $i$the object in the data set that
are contained in the subtree rooted at $v$.
We apply $k$-means clustering to these vectors and, finally, assign weights to
each cluster. Therefore, all nodes in the same cluster share the same weight
learned by the MKL algorithm.

\section{Experimental Evaluation}

We performed classification experiments using the $C$-SVM implementation 
LIBSVM and the EasyMKL implementation of MKLpy v0.2.1b0.\footnote{\url{https://pypi.org/project/MKLpy/}}
We report average prediction accuracies and standard deviations obtained by 
$5$-fold cross-validation repeated $5$ times with random fold assignment.
Within each fold all hyperparameters were selected by cross-validation 
based on the training set. The regularization parameter $C$ was selected from
$\{0.01, 0.1, 1, 10, 100\}$ and the parameter $\lambda$ from 
$\{0.1, 0.3, 0.5, 0.7, 0.9\}$.

We compare the Weisfeiler-Lehman subtree kernel (WL), the Weisfeiler-Lehman
optimal assignment kernel (WL-OA) and its deep variant without feature grouping
(DWL-OA1) and with feature grouping (DWL-OA2), where the number of clusters
was set to $k=10$.
We set the number of color refinement iterations to $h=4$ for all experiments.
We tested on widely-used graph classification benchmarks from different domains~\cite{KKMMN2016}.
\textsc{Mutag}, \textsc{PTC-MR}, \textsc{NCI1} and \textsc{NCI109} are graphs 
derived from small molecules, \textsc{Proteins} and \textsc{D\&D} represent 
macromolecules, and \textsc{Reddit} contains social network graphs.
All data sets consist of two classes, the vertex and edge labels were ignored, if 
present.

\setlength\tabcolsep{3.7pt}
\newcommand{\win}[1]{$\hspace{-0.3mm}$\textbf{#1}}
\newcommand{\sd}[1]{\scriptsize{$\pm$#1}}
\begin{table*}\small
\begin{center}
  \caption{Classification accuracies and standard deviations on graph data sets 
  representing small molecules, macromolecules and social networks.}
  \label{tab:accuracies}
  \begin{tabular}{@{}lccccccc@{}}\toprule
    \multirow{3}{*}{\textbf{Kernel}}    &\multicolumn{7}{c}{\textbf{Data Set}}\\\cmidrule{2-8}
                     & \textsc{Mutag}   & \textsc{PTC-MR}  & \textsc{NCI1}    & \textsc{NCI109}  & \textsc{Proteins} & \textsc{D\&D}    &  \textsc{Reddit}  \\\midrule
    \textsc{WL}      & 88.3\sd{1.2}     & 54.5\sd{2.5}     & 78.9\sd{0.6}     & 79.6\sd{0.3}     & 70.6\sd{0.5}      & 73.2\sd{0.4}     & 71.7\sd{0.3}      \\
    \textsc{WL-OA}   & 88.6\sd{1.0}     & 55.8\sd{2.3}     & 78.6\sd{0.3}     & 78.8\sd{0.6}     & 73.8\sd{0.4}      & 75.7\sd{0.3}     & 88.5\sd{0.3}      \\
    \textsc{DWL-OA1} & 88.1\sd{1.0}     & 56.7\sd{2.3}     & OOM              & OOM              & OOM               & OOM              & OOM               \\
    \textsc{DWL-OA2} & 88.5\sd{0.9}     & 57.6\sd{1.3}     & 78.9\sd{0.2}     & 78.5\sd{0.2}     & 72.5\sd{0.4}      & 76.7\sd{0.3}     & 86.9\sd{0.2}      \\
    \bottomrule
  \end{tabular}
\end{center}
\end{table*}

We were not able to  run DWL-OA1 on the large date sets with more 
than thousand objects due to memory constraints.
All results are summarized in Table~\ref{tab:accuracies}. 
We observed only minor differences in accuracy between the three Weisfeiler-Lehman
optimal assignment kernels. 
DWL-OA2 performs better than DWL-OA1 which indicates 
the benefit of feature grouping.
For the considered data sets, DWL-OA has obtained state-of-the-art accuracy 
results, but there is no clear evidence that learning weights via MKL improves the 
classification accuracy significantly.
However, we have observed that a significant proportion of the learned weights is 
zero, which leads to compact sparse models.

\section{Conclusion}
We have proposed Weisfeiler-Lehman assignment kernels which learn deep representations
for graph classification.
It remains future work to analyze the learned weights and their domain-specific 
meaning in detail. We believe that the interpretability of the weights
in terms of vertex neighborhoods is a strength of the approach and can give
new insights into real-world problems.
Our method only allows to learn vertex similarities from a predefined
restricted class of functions. In the future, we would like to study more general 
approaches such as learning the entire hierarchy and not just its weights.

\bibliographystyle{unsrt}
\bibliography{lit.bib}

\end{document}